\documentclass[letterpaper, 10pt, conference]{ieeeconf}      %

\IEEEoverridecommandlockouts                              %

\usepackage{hyperref} 
\usepackage{mathrsfs}
\usepackage{amsfonts}
\usepackage{amsmath}
\usepackage[flushleft]{threeparttable}
\usepackage{tablefootnote}
\usepackage{multirow}
\usepackage{graphicx}
\usepackage{subfigure}
\let\labelindent\relax
\usepackage{enumitem}
\usepackage{hanging}
\usepackage{color}
\usepackage{tikz}
\usetikzlibrary{calc,arrows,decorations.markings}
\usepackage{balance}
\usepackage{tabularx}

\newcommand{\R}{\mathbb{R}}  %

\graphicspath{{image}}

\usepackage{booktabs}  %

\title{\LARGE \bf
Keypoint-Based Category-Level Object Pose Tracking from an RGB Sequence with Uncertainty Estimation}

\author{Yunzhi Lin$^{1,2}$, Jonathan Tremblay$^{1}$, Stephen Tyree$^{1}$, Patricio A.  Vela$^{2}$, 
Stan Birchfield$^{1}$ 
\\ $^{1}$NVIDIA: {\tt\small \{jtremblay, styree, sbirchfield\}@nvidia.com}
\\ $^{2}$Georgia Institute of Technology: {\tt\small \{yunzhi.lin, pvela\}@gatech.edu}
\thanks{Project: \href{https://sites.google.com/view/centerposetrack}{sites.google.com/view/centerposetrack}} \thanks{Work was completed while the first author was an intern at NVIDIA.}%
\thanks{This work was supported in part by NSF Award \#2026611.}%
}

\begin{document}

\maketitle
\thispagestyle{empty}
\pagestyle{empty}

\begin{abstract}
We propose a single-stage, category-level 6-DoF pose estimation algorithm that 
simultaneously \emph{detects} and \emph{tracks} instances of objects within a known category.
Our method takes as input the previous and current frame from a monocular RGB video, as well as predictions from the previous frame, to predict the bounding cuboid and 6-DoF pose (up to scale).
Internally, a deep network predicts distributions over object keypoints (vertices of the bounding cuboid) in image coordinates, after which a novel probabilistic filtering process integrates across estimates before computing the final pose using P$n$P.
Our framework allows the system to take  
previous uncertainties into consideration when predicting the current frame, resulting in predictions that are more accurate and stable than single frame methods. 
Extensive experiments show that our method outperforms existing approaches on the challenging Objectron benchmark of annotated object videos.
We also demonstrate the usability of our work in an augmented reality setting.

\end{abstract}

\section{Introduction}

In robotics and augmented reality settings, detecting object poses in three dimensions is crucial.
Once we start interacting with an environment, objects that have been detected also need to be tracked. 
Pose tracking encourages temporally consistent pose predictions, allowing past observations to inform current predictions.
For robotic manipulation, pose tracking can bring robustness to semantically meaningful interaction over time, or aid in keeping virtual worlds in sync with the robot environment.

The problem of 6-DoF pose tracking is a rich topic in the computer vision community.
Recent object pose tracking methods~\cite{pauwels2015simtrack, tjaden2017real} have focused on the instance-level problem, 
which assumes the 3D model of the target instance is available. 
Template matching~\cite{li2018deepim, wen2020se} is among the most popular methods. 
It aims to compute the relative pose between two images by comparing a rendered synthetic image from the previous pose estimation with the current realistic input. 
By harnessing the power of rendering techniques, these methods can achieve remarkable accuracy and robustness. 
However, the assumption of the known 3D model 
is not always valid in realistic settings, where high-fidelity 3D models of novel objects are costly to acquire.

In this work, we seek to extend reliable 6-DoF pose tracking to categories of objects, 
while using input from only a single RGB camera and no 3D instance meshes.
We consider a category to be a class of objects of similar physical shapes, {\em e.g.}, cereal boxes, shoes, or mugs.
Category-level pose prediction presents an additional challenge to the instance-level pose problem: whereas pose prediction for known instances encompasses translation and rotation, for categories of objects with varying size and shape we also need to predict and track the object dimensions (\emph{i.e.,} width, height, and length).

\begin{figure}[t!]
  \centering
    \includegraphics[width=\columnwidth]{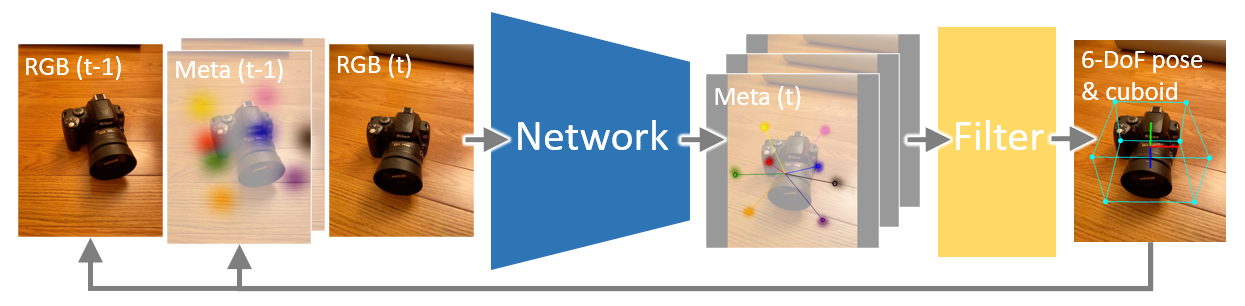}
  \caption{Overview of our CenterPoseTrack framework for category-level 6-DoF object pose tracking from a monocular RGB sequence. Our method leverages a deep network and a filtering process to estimate and propagate 2D projected keypoints of 3D bounding box vertices, using an off-the-shelf P$n$P process to compute the final 6-DoF pose and relative dimensions. 
  Please note that for simplicity `meta' is used to refer to 
  keypoints, bounding boxes, sub-pixel offsets, uncertainties, {\em etc.}} 
  \label{fig:abstract}
  \vspace*{-3 ex}
\end{figure}

While the research community has given much attention to the category-level 
pose estimation problem~\cite{wang2019normalized,tian2020shape,chen2020learning,chen2020category,manhardt2020cps++}, 
few works have considered category-level tracking.
The most relevant work to ours is 6-PACK~\cite{wang20206}, 
which takes as input a single RGB-D frame (color + depth) from which anchor points are generated in an unsupervised manner.
Using the anchor points, the algorithm tracks the object over time by matching the points from the current frame with the previous one. 
Another work, CAPTRA~\cite{weng2021captra}, uses point clouds to track object positions over time. 
In contrast to these works, we focus exclusively on leveraging the accessibility of RGB-only inputs, 
while also integrating uncertainties which current category-level pose tracking methods have not fully explored.

The emerging trend of jointly performing detection and tracking~\cite{feichtenhofer2017detect, zhang2018integrated, bergmann2019tracking} has made progress in overcoming possible issues with bad predictions coming from either current or previous frames. 
Inspired by these successes and our previous work~\cite{lin2021arx:centerpose} on category-level pose estimation, we propose a keypoint-based solution to track objects in three dimensions. We represent an object as a set of vertex keypoints from its 3D cuboid projected onto the image plane. This representation allows us to leverage P$n$P for computing the final object pose.
Figure~\ref{fig:abstract} presents an overview of our framework: Our tracker is composed of two parts: a deep neural network and a filtering algorithm.
The deep network takes as input the current observation, {\em i.e.}, the image at time $t$,
as well as the observation and high-level predictions from time $t-1$. 
High-level outputs from the network are then sent to our filtering process. 
We first compute a fused distribution over
predicted keypoints using Bayesian filtering on the network outputs, followed by a Kalman filter to propagate these predictions in time.
In order to keep the object size consistent, we leverage Bayesian filtering over all the previous cuboid size predictions.
With that information, the final object pose is computed via P$n$P. 
Using the uncertainty estimates returned by the Kalman filter, we render heatmap inputs for the next frame, 
allowing the network to reason about the likelihood of keypoint positions.

Our work makes the following contributions:
\begin{itemize}
    \item We propose a category-level method called CenterPoseTrack for detecting and tracking 6-DoF object poses and dimensions (up to scale) using monocular RGB video as input. 
    \item We show the importance of incorporating uncertainty estimation, which is performed through both a tracklet-conditioned deep network and a probabilistic filtering process. 
    \item The evaluation of the proposed method demonstrates improved generalization and robustness on the Objectron benchmark~\cite{ahmadyan2021objectron}. %
\end{itemize}

\section{Related Work}

{\bf 6-DoF object pose estimation.} 
6-DoF object pose estimation infers the 3D translation and rotation of a target object. State-of-the-art methods can be generally categorized into template matching~\cite{zeng2017multi,sundermeyer2018implicit} and regression techniques~\cite{xiang2018posecnn,rad2017bb8,tremblay2018deep,peng2019pvnet}. While these approaches have achieved impressive results, they rely on the availability of an instance-specific 3D CAD model of the object at training, annotation, and/or inference time. 
In order to extend this body of work, recent 
methods have explored the category-level pose estimation, 
finding the 6-DoF pose of a previously unseen object selected from a known object category, {\em e.g.}, mugs or shoes.
In this context, prior work has
explored learning a normalized object coordinate space~\cite{wang2019normalized}, 
matching pose-dependent and pose-independent features separately~\cite{chen2020learning}, 
or
modeling deformation from the categorical shape prior~\cite{tian2020shape}. 
Most of these methods leverage large collections of synthetic 3D CAD models to generate complex annotations of rendered images during training time.
An alternative approach that is similar to our work directly regresses the 2D projections object keypoints \cite{hou2020mobilepose, ahmadyan2021objectron},
using large-scale real-world image datasets annotated with 3D object bounding box \cite{ahmadyan2021objectron}. 
Our prior work \cite{lin2021arx:centerpose} belongs to this category and lays the foundation of our proposed tracking framework. 
Although our previous detector has achieved state-of-art performance on the Objectron dataset~\cite{ahmadyan2021objectron}, 
we show in this work that it does not keep temporal consistency across frames.

{\bf 6-DoF object pose tracking.}
A natural extension to single frame pose estimation is tracking objects over time. 
Current instance-level tracking approaches can be divided into two groups: probabilistic- and optimization-based methods.
The former methods build frameworks on different filtering strategies, including particle filtering~\cite{choi2013rgb, wuthrich2013probabilistic,deng2019pose} or Gaussian filtering~\cite{issac2016depth}. 
The latter aim to capture the discrepancy between the current observation and the previous state, then computes inter-frame change in pose by minimizing the residual function in a least-squares manner ~\cite{pauwels2015simtrack, joseph2015versatile, tjaden2017real, li2018deepim, wen2020se}. 
When extending tracking to the category-level pose problem, the task becomes more challenging as features cannot not be extracted from known CAD models. 
The pioneering work of 6-PACK~\cite{wang20206} proposed a novel anchor-based keypoint generation neural network that reliably detects the same keypoints from the point cloud and uses them to estimate the inter-frame change in pose through keypoint matching.
CAPTRA~\cite{weng2021captra} adopts a per-part cannibalization module for point clouds, then uses two separate networks to estimate the change of each part's coordinates and rotation between frames. While these depth-based methods are successful, category-level 6-DoF pose tracking problem from monocular RGB input has not been fully explored. 

{\bf Uncertainty modeling.}
Prediction uncertainty may be decomposed into epistemic and aleatoric uncertainty~\cite{kendall2017uncertainties,feng2020review}. Epistemic uncertainty arises in the model parameters which reflect a limited set of training data, while
aleatoric uncertainty stems from sensor error or label noise. 
Many efforts have been put in estimating the uncertainty of 2D or 3D bounding boxes in object detection~\cite{pan2020towards, he2020deep, harakeh2020bayesod, dong2020probabilistic}. 
Most of them are based on the idea of direct modeling, which assumes a particular probability distribution over network outputs, and uses additional layers to estimate the parameters of the hypothesized distribution.
However, its importance has not been fully realized for the category-level pose tracking problem, where the intra-class shape variability within a category remains a key challenge~\cite{lin2021arx:centerpose}. 
In this work, we leverage recent progress~\cite{meyer2020learning} on modeling aleatoric uncertainty and apply it to predicting inter-frame keypoint displacement and relative cuboid dimensions. These uncertainty estimates serve as a crucial component for prediction fusion in the downstream pipeline.

\begin{figure*}[h]
  \centering
    \includegraphics[width=\textwidth]{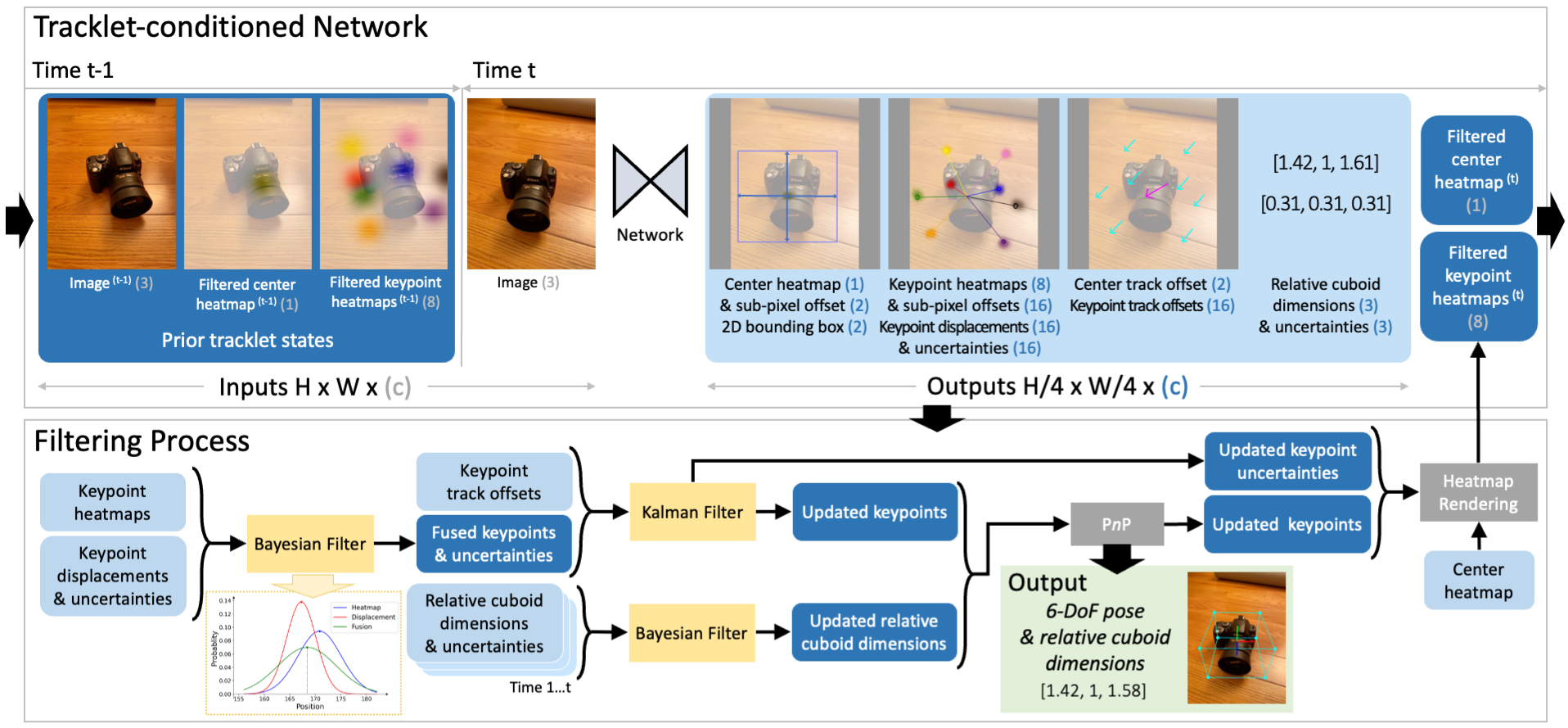}
  \caption{Overview of our CenterPoseTrack method, which consists of two parts. {\sc Top:} Tracklet-conditioned network: At time $t$, the network takes as input the current image frame $I^{t} \in \R^{H \times W \times 3}$, along with the previous frame $I^{t-1}$ and rendered heatmaps from the prior tracklet at time $t-1$ (dark blue). We employ the CenterPose method~\cite{lin2021arx:centerpose} as the network.  \,\,\, {\sc Bottom:} Filtering process: Given the network predictions (light blue), we apply filters to update the 2D keypoint locations as well as the relative cuboid dimensions. The obtained result is further processed by P$n$P to get the final 6-DoF pose. Finally, predictions are rendered on new heatmaps as inputs for the next time step.
    \label{fig:pipeline}}
  \vspace*{-3 ex}
\end{figure*}

\section{Approach}

Inspired by CenterPose~\cite{lin2021arx:centerpose}, we developed a similar architecture that 
allows tracking of objects.
Our approach to category-level object pose tracking is illustrated in greater detail in Figure~\ref{fig:pipeline}.
At a high level, the upper section describes the inputs and outputs of the trained neural network.
The lower section depicts  %
the filtering process that reasons over network outputs and produces a final pose estimate.
This section presents the components used by CenterPoseTrack.

\subsection{Background}

We assume as input  
an RGB image of the current frame $I^{(t)} \in \mathbb{R}^{H \times W \times 3}$
and the previous frame $I^{(t-1)} \in \mathbb{R}^{H \times W \times 3}$, as well as the predictions for the tracked objects in the previous frame $T^{(t-1)}=\left\{p_{0}^{(t-1)}, p_{1}^{(t-1)}, \ldots\right\}$. 
More specifically, each tracklet $p$ is described by its
object center ($\mathbf{p_{cen}} \in \mathbb{R}^{2}$), 
2D bounding box ($\mathbf{s} \in \mathbb{R}^{2}$), 
8 corner keypoints ($\mathbf{p_{key}} \in \mathbb{R}^{2}$) with uncertainties ($\mathbf{\sigma_{key}} \in \mathbb{R}^{2}$), 
relative dimensions ($\mathbf{d} \in \mathbb{R}^{3}$) with uncertainties ($\mathbf{\sigma_{dim}} \in \mathbb{R}^{3}$), and a unique identity ({\em id} $\in \mathbb{Z}^+$).
The goal is to detect and track objects $T^{(t)}=\left\{p_{0}^{(t)}, p_{1}^{(t)}, \ldots\right\}$ in the current frame $I^{(t)}$, as well as assigning objects with a consistent {\em id}.
If there are no previously found objects, then the method is equivalent to detection.

\subsection{Tracklet-conditioned detection \label{sec:approach_condition}}

A natural way to increasing temporal coherence is to provide the detector with additional image inputs from past frames.
Considering the additional computation cost brought by adding more images, we only include a single previous frame~\cite{zhou2020tracking}.
We also render heatmaps (object center and keypoints) from detections in the previous frame $I^{(t-1)}$
as inputs to the neural network. 
In Figure~\ref{fig:pipeline}, these inputs are referred as {\em Filtered center heatmap} and {\em Filtered keypoint heatmaps}.

We leverage rendering to include uncertainty from the previous frame's prediction.
Specifically we use a scaled Gaussian distribution, where the scale $k= \max \left(1-\mathrm{c}^{\frac{a- \mathbf{\sigma}}{a-b}}, 0\right)$, and $\sigma$ is the corresponding predicted uncertainty (we use $a=9, b=3,$ and $c=0.15$).

The network is also trained to output 2D offset vectors from the previous keypoint locations to current locations. %
This offset prediction, similar to sparse optical flow, is an efficient way to link points across different frames~\cite{zhou2020tracking}. 
In our experiments, we found that the network can estimate the change in the scene and potentially recover occluded keypoints at time $t$ from visual evidence at time $t-1$.
With the proposed design, the network is able to implicitly reason over prior detections with temporal coherency.

\subsection{Training data generation \label{sec:approach_data}}

The main challenge in training such a network comes from generating realistic heatmap inputs as training examples. 
At inference time, 
the heatmaps can contain an arbitrary number of missing keypoints, 
predictions with localization error, 
or even false positives. 
We employ two modes for data generation:
1) Using predictions from CenterPose \cite{lin2021arx:centerpose}; 
2) Modeling the test-time error at training time, 
inspired by~\cite{tremblay2018synthetically,zhou2020tracking}.
For the latter, three types of noise are applied to the ground truth annotations for the object center and the corner keypoints.
Gaussian noise $n \sim {\cal N}(0,\sigma^2)$ is added to the locations. 
False positives are added by rendering on the heatmap new points with probability $\lambda_{f p}$.
False negatives also remove randomly points with probability $\lambda_{f n}$. 
In our final implementation, we set $\sigma=1$, $\lambda_{f p}=0.1$, and $\lambda_{f n}=0.2$ for the object center while $\sigma=1$, $\lambda_{f p}=0.05$, and $\lambda_{f n}=0.1$ for other keypoints. 
The Gaussian noise is rendered based on a scale determined by the noise level, where the scale $k= \max \left(1-\alpha^{n-\beta}, 0\right)$, $\alpha=2$ and $\beta=4.5$. 
During noise simulation, 
keypoints are rendered only if their corresponding object center has already been rendered. 
We also %
randomly sample from all frames $I^{(k)}$ where $|k-t|<M_{f}$ near the current frame $I^{(t)}$ to avoid over-fitting to the frame rate.

\subsection{Uncertainty Modeling}

Capturing uncertainties
(due to perception inaccuracy or sensor noise)
provides valuable information for the tracking pipeline.
When directly predicting uncertainty as a network output, it is common practice to impose a negative log likelihood loss function~\cite{kendall2017uncertainties, kraus2019uncertainty}:
\begin{equation}
\mathcal{L}_{NLL}(y, \hat{y}, \hat{b})=\log 2 \hat{b}+\frac{|y-\hat{y}|}{\hat{b}},
\label{equ:null}
\end{equation}
where $y$ refers to the ground truth label, $\hat{y}$ is the predicted output, 
and $\hat{b}$ represents the predicted uncertainty.
In contrast, we introduce a  Kullback-Leibler (KL) divergence loss function inspired by~\cite{meyer2020learning} 
for the displacement and relative dimension outputs:
\begin{equation}
\mathcal{L}_{KLD}(y, \hat{y}, \hat{b}, \beta)=\log \frac{\hat{b}}{\beta}+\frac{\beta \exp \left(-\frac{|y-\hat{y}|}{\beta}\right)+|y-\hat{y}|}{\hat{b}}-1,
\label{equ:kld}
\end{equation}
with label uncertainty hyperparameter $\beta$.
The KL divergence loss has a two-fold advantage over negative log likelihood.
First, the KL divergence is always greater than 0 while the negative log likelihood goes to negative infinity as the uncertainty $\rightarrow 0$ and error $\rightarrow 0$. 
Using the KL divergence can prevent this individual loss term from dominating the overall loss.
Second, the shape of the KL divergence can easily be controlled by the hyperparameter $\beta$,
which is useful for preventing the model from overfitting to noisy labels.

For better convergence, %
we modify Eq.~\eqref{equ:kld} to:
\begin{equation}
\begin{aligned}
&\mathcal{L}_{KLD+}(y, \hat{y}, {\hat{\sigma}}^{2}, \beta^{2})= \\
&\log \frac{\lambda \hat{\sigma}^{2}}{\beta^{2}}+\frac{\beta^{2} \exp \left(-\frac{(y-\hat{y})^{2}}{\beta^{2}}\right)+(y-\hat{y})^{2}}{\lambda\hat{\sigma}^{2}}-1+\frac{1}{2} \lambda \hat{\sigma}^{2},
\end{aligned}
\label{equ:kld+}
\end{equation}
where ${\hat{\sigma}}^{2}$ represents the predicted uncertainty and $\beta^{2}$ represents the label uncertainty hyperparameter 
(set to 0.1 in our experiments). 
We predict $\log \lambda \hat{\sigma}^{2}$ instead of ${\hat{\sigma}}^{2}$ and adopt a compulsory gradient clipping step for the overall loss (bound is set to 100) for better stability during the training process. 
We also use $\frac{1}{2} \lambda \hat{\sigma}^{2}$ as an additional regularization term. 
Finally, $\lambda$ is used as a coefficient hyperparameter which is set to 0.25.

\subsection{Filtering Process \label{sec:approach_filter}} 

Tracking methods have to answer the fundamental question of how much confidence to have in the prior predictions {\em vs.} 
the current ones. 
In this work, we have two ways to model this trade-off. 
In Section \ref{sec:approach_condition}, we have already illustrated how we incorporate 
the probabilistic heatmap rendering as a network input.
However, the network only has access to a local window (2 frames), 
while information in the longer horizon may be essential for accurate and stable tracking.

In order to leverage long horizon information (see bottom of Fig.~\ref{fig:pipeline}),
we first fuse the keypoint heatmap (the heat center) and the displacement prediction inputs via a Bayesian filter~\cite{kumar2006generalized}, where input number $n = 2$ in our case and $\hat{\mu}_{i}$ represents the corresponding 2D location prediction:
\begin{equation}
\hat{\sigma}=\left(\sum_{i=1}^{n} \hat{\sigma}_{i}^{-2}\right)^{-1 / 2}, \qquad \hat{\mu}=\hat{\sigma}^{2} \sum_{i=1}^{n} \hat{\sigma}_{i}^{-2} \hat{\mu}_{i}.
\label{equ:bauesian}
\end{equation}
Then, the updated keypoints with corresponding uncertainties, and the predicted keypoint offsets, are fed into a Kalman filter as observable variables.
This allows us to represent locations, uncertainties for the location estimates, and the object velocities.
A simple constant velocity model is used, and the uncertainties for the velocity estimates are set at 20 in our experiments.
As the relative dimension prediction inputs stay the same for one specific object over time, we employ another Bayesian filter~\cite{kumar2006generalized} on the entire history of their measurements.
After the updates, the keypoint estimates from the Kalman filter along with the estimated relative dimensions, 
are fed to a Levenberg-Marquardt version of P$n$P~\cite{abdel2015direct}, 
producing the final 6-DOF pose output. 
Using the projected keypoints with uncertainties, and the likelihood for the center predictions, we render new heatmaps (Section \ref{sec:approach_condition}) for the next timestep.

\subsection{Loss Function}

We adopt the focal loss for center $\mathcal{L}_{\text {p}_{cen}}$ and keypoint heatmaps $\mathcal{L}_{\text {p}_{key}}$.  
We adopt the L1 loss for 2D bounding box size $\mathcal{L}_{\text {bbox}}$, center offset $\mathcal{L}_{\text {off}}$, center tracking offset $\mathcal{L}_{\text track}$, keypoint offset $\mathcal{L}_{\text {off}_{key}}$, and keypoint tracking offset $\mathcal{L}_{\text {track}_{key}}$. We employ our proposed KLD+ loss for displacement $\mathcal{L}_{\text {KLD+}_{dis}}$ and relative dimension $\mathcal{L}_{\text {KLD+}_{dim}}$. The overall training objective is the weighted combination of these nine terms.

\subsection{Implementation Details} 

Our network was trained with a batch size of 32 on 4 NVIDIA V-100 GPUs for 15 epochs, starting with the pretrained ImageNet weights. We use the Adam optimizer with an initial learning rate of 2.5e-4, dropping 10x at both 6 and 10 epochs.
On average, around 60~h are required to train one category (using between 120k to 500k training images depending on the category). 
Inference speed is around 8 fps on an NVIDIA GTX 1080Ti GPU.

\section{Experimental Results \label{sec:results}}

In this section, we aim to demonstrate that our tracking algorithm achieves state-of-the-art results on the Objectron dataset~\cite{ahmadyan2021objectron} and that it alleviates the stability issues seen with previous detection work~\cite{hou2020mobilepose, lin2021arx:centerpose} when predicting over multiple frames. 
We also demonstrate the use of CenterPoseTrack in a real-world augmented reality system.

\subsection{Dataset}

We evaluate our method on the Objectron dataset~\cite{ahmadyan2021objectron} which contains 9 categories: bikes, books, bottles, cameras, cereal boxes, chairs, cups, laptops, and shoes. There are 15k video clips with over 4M frames annotated with 3D object bounding boxes, camera poses, sparse point clouds, and surface planes.
Categories with symmetric objects are handled following~\cite{wang2019normalized,lin2021arx:centerpose}.

\subsection{Metrics}

Following~\cite{lin2021arx:centerpose}, we report the following metrics: 1) average precision of 3D intersection over union (IoU) at 50$\%$, 2) mean pixel error of the 2D projection of cuboid vertices, 3) average precision of azimuth at $15^{\circ}$, and 4) average precision of elevation at $10^{\circ}$. 
For the symmetric object categories (bottle$^*$ and cup$^*$), we maximize 3D IoU or minimize 2D pixel projection error by rotating the predicted bounding box along the symmetry axis N times ($N=100$ following \cite{ahmadyan2021objectron}).

We also introduce a new consistency metric for tracking of static objects,
aiming to measure how much a prediction changes in translation, rotation, and cuboid dimensions over time.
We propose to compare multi-frame predictions against each other using 3D IoU in the world coordinate frame.
A perfectly consistent tracking system would have perfect overlap between pairs of predictions, {\em e.g.}, 3D IoU of~$1.0$.
Since $n^2$ comparisons can be computationally intensive, where {\em n} is the number of frames, 
we report the average of consistency score computed within a 5-frame sliding window.

\begin{table*}[h]
 \vspace*{0.06in}
\caption{Pose estimation results on the Objectron test set~\cite{ahmadyan2021objectron}. \\
Note that our CenterPoseTrack outperforms competing methods on all 3D metrics.
}
\label{tab:comparison_combined}
  \centering
\begin{tabular}{cccccccccccc}
\toprule
      Method     & Bike   & Book   & Bottle$^{*}$ & Camera & Cereal\_box & Chair  & Cup$^{*}$ & Laptop & Shoe   & Mean   \\ 
\midrule
\multicolumn{11}{c}{Average precision at 0.5 3D IoU ($\uparrow$)} \\
\cmidrule(r){4-6}
MobilePose \cite{hou2020mobilepose} & 0.3109 & 0.1797 & 0.5433 & 0.4483 & 0.5419 & 0.6847 & 0.3665 & 0.5225 & 0.4171 & 0.4461 \\
Two-stage \cite{ahmadyan2021objectron} & 0.6127          & 0.5218          & 0.5744          & \textbf{0.8016} & 0.6272          & 0.8505          & 0.5388          & 0.6735          & 0.6606          & 0.6512           \\ 
CenterPose~\cite{lin2021arx:centerpose}        & 0.6419 & 0.5565 & 0.8021 & 0.7188 & 0.8211 & 0.8471 & 0.7704 & 0.6766 & 0.6618 & 0.7218       \\
CenterPose~\cite{lin2021arx:centerpose}  w/ filtering      & 0.6176          & 0.5757          & 0.7967          & 0.7335          & 0.8347          & 0.8500          & 0.7263          & 0.6533          & 0.6827          & 0.7189          \\
Ours w/o filtering   & 0.6820          & 0.6023          & 0.7544          & 0.7475          & 0.8378          & 0.8604          & 0.8534          & 0.7139          & 0.6278          & 0.7422          \\
Ours w/o heatmap        & 0.6846          & 0.7064          & 0.6540          & 0.7648          & 0.8405          & 0.8707          & 0.6900          & 0.6550          & 0.5954          & 0.7179          \\
Ours       & \textbf{0.7389} & \textbf{0.7829} & \textbf{0.8256} & 0.7835          & \textbf{0.8598} & \textbf{0.8927} & \textbf{0.8991} & \textbf{0.7344} & \textbf{0.7274} & \textbf{0.8049}       \\
\\
\multicolumn{11}{c}{Mean consistency score on 3D IoU ($\uparrow$)} \\
\cmidrule(r){4-6}
CenterPose~\cite{lin2021arx:centerpose}        & 0.8760          & 0.8501          & 0.8833          & 0.8888          & 0.9056          & 0.9138          & 0.8906          & 0.7744          & 0.8881          & 0.8745          \\
CenterPose~\cite{lin2021arx:centerpose}  w/ filtering      & 0.8777          & 0.8612          & 0.8923          & 0.8979          & 0.9109          & 0.9119          & 0.8668          & 0.7792          & 0.8978          & 0.8773          \\
Ours w/o filtering   & 0.8881          & 0.8685          & 0.8798          & 0.8919          & 0.9089          & 0.9221          & 0.9066          & 0.7916          & 0.8908          & 0.8832          \\
Ours w/o heatmap        & \textbf{0.9172} & 0.8710          & \textbf{0.8999} & \textbf{0.9098} & 0.9181          & 0.9311          & 0.8641          & 0.7996          & 0.8895          & 0.8889          \\
Ours       & 0.9150          & \textbf{0.8942} & 0.8940          & 0.9088          & \textbf{0.9214} & \textbf{0.9319} & \textbf{0.9160} & \textbf{0.8064} & \textbf{0.9043} & \textbf{0.8991} \\
\\
\multicolumn{11}{c}{Mean pixel error of 2D projection of cuboid vertices ($\downarrow$)} \\
\cmidrule(r){4-6}
MobilePose \cite{hou2020mobilepose} & 0.1581 & 0.0840 & 0.0818 & 0.0773 & 0.0454 & 0.0892 & 0.2263 & 0.0736 & 0.0655 & 0.1001\\
Two-stage  \cite{ahmadyan2021objectron}  & 0.0828          & \textbf{0.0477} & 0.0405          & \textbf{0.0449} & 0.0337          & \textbf{0.0488} & 0.0541          & \textbf{0.0291} & \textbf{0.0391} & 0.0467          \\
CenterPose~\cite{lin2021arx:centerpose}        & 0.0872          & 0.0563          & \textbf{0.0400} & 0.0511          & 0.0379          & 0.0594          & 0.0376          & 0.0522          & 0.0463          & 0.0520          \\
CenterPose~\cite{lin2021arx:centerpose}  w/ filtering      & 0.0877          & 0.0560          & 0.0403          & 0.0513          & 0.0378          & 0.0621          & 0.0420          & 0.0594          & 0.0455          & 0.0536          \\
Ours w/o filtering   & 0.0792          & 0.0514          & 0.0447          & 0.0507          & 0.0334          & 0.0545          & 0.0387          & 0.0482          & 0.0459          & 0.0496          \\
Ours w/o heatmap        & 0.0767          & 0.0614          & 0.0432          & 0.0493          & 0.0345          & 0.0506          & 0.0474          & 0.0602          & 0.0486          & 0.0524          \\
Ours       & \textbf{0.0748} & 0.0502          & 0.0413          & 0.0491          & \textbf{0.0316} & 0.0509          & \textbf{0.0359} & 0.0400          & 0.0427          & \textbf{0.0463} \\
\\
\multicolumn{11}{c}{Average precision at $15^{\circ}$ azimuth error ($\uparrow$)} \\
\cmidrule(r){4-6}
MobilePose \cite{hou2020mobilepose} & 0.4376          & 0.4111          & 0.4413          & 0.5293          & 0.8780          & 0.6195          & 0.0893          & 0.6052          & 0.3934          & 0.4894          \\
Two-stage  \cite{ahmadyan2021objectron}  & 0.8234          & 0.7222          & 0.8003          & 0.8030          & 0.9404          & 0.8840          & 0.6444          & 0.8561          & 0.5860          & 0.7844          \\
CenterPose~\cite{lin2021arx:centerpose}       &0.8622          & 0.7323          & 0.9561          & 0.8226          & 0.9361          & 0.8822          & 0.8945          & 0.7966          & 0.6757          & 0.8398          \\
CenterPose~\cite{lin2021arx:centerpose}  w/ filtering      & 0.8516          & 0.7316          & \textbf{0.9652} & \textbf{0.8335} & 0.9351          & 0.8839          & 0.8947          & 0.7841          & \textbf{0.6764} & 0.8396          \\
Ours w/o filtering   & \textbf{0.8814} & 0.8001          & 0.9501          & 0.8333          & 0.9496          & 0.8927          & 0.9680          & 0.8648          & 0.6465          & \textbf{0.8652}          \\
Ours w/o heatmap        & 0.8479          & 0.8041          & 0.8874          & 0.8129          & 0.9352          & 0.8806          & 0.8724          & 0.8146          & 0.6730          & 0.8365          \\
Ours       & 0.8525          & \textbf{0.8182} & 0.9411          & 0.8276          & \textbf{0.9497} & \textbf{0.8940} & \textbf{0.9687} & \textbf{0.8659} & 0.6632          & 0.8645     \\
\\
\multicolumn{11}{c}{Average precision at $10^{\circ}$ elevation error ($\uparrow$)} \\
\cmidrule(r){4-6}
MobilePose \cite{hou2020mobilepose} & 0.7130          & 0.6289          & 0.6999          & 0.5233          & 0.8030          & 0.7053          & 0.6632          & 0.5413          & 0.4947          & 0.6414          \\
Two-stage  \cite{ahmadyan2021objectron}  & \textbf{0.9390} & 0.8616          & 0.8567          & 0.8437          & 0.9476          & \textbf{0.9272} & 0.8365          & 0.7593          & 0.7544          & 0.8584          \\
CenterPose~\cite{lin2021arx:centerpose}        & 0.9072          & 0.8535          & 0.8881          & \textbf{0.8704} & 0.9467          & 0.8999          & 0.8562          & 0.6922          & \textbf{0.7900} & 0.8560          \\
CenterPose~\cite{lin2021arx:centerpose}  w/ filtering      & 0.8988          & 0.8412          & 0.8688          & 0.8477          & 0.9458          & 0.9066          & 0.8061          & 0.6765          & 0.7864          & 0.8420          \\
Ours w/o filtering   & 0.9134          & \textbf{0.8978} & 0.8984          & 0.8439          & 0.9516          & 0.9104          & 0.9372          & 0.7554          & 0.7549          & \textbf{0.8737}          \\
Ours w/o heatmap        & 0.8906          & 0.8776          & 0.8461          & 0.8126          & 0.9526          & 0.8915          & 0.7858          & 0.7260          & 0.7355          & 0.8354          \\
Ours       & 0.9029          & 0.8976          & \textbf{0.9014} & 0.8296          & \textbf{0.9561} & 0.9111          & \textbf{0.9422} & \textbf{0.7620} & 0.7376          & 0.8712
\\
\bottomrule
\label{table:main}
\end{tabular}
\end{table*}

\begin{table*}[t]
\caption{Effect of track initialization on pose estimation (average precision at 0.5 3D IoU metric ($\uparrow$)).  \\
Note that our CenterPoseTrack does not need initialization to perform well.
\label{tab:init}}
  \centering
\begin{tabular}{ccccccccccc}
\toprule
     Initialization & Bike   & Book   & Bottle$^{*}$ & Camera & Cereal\_box & Chair  & Cup$^{*}$ & Laptop & Shoe   & Mean   \\
\midrule
GT  & 0.7389 & 0.7829 & 0.8256 & 0.7835 & 0.8598 & 0.8927 & 0.8991 & 0.7344 & 0.7274 & 0.8049 \\ 
Noisy GT & 0.7167 & 0.6911 & 0.7913 & 0.7589 & 0.7845 & 0.8916 & 0.8874 & 0.7316 & 0.7078 & 0.7734 \\

CenterPose~\cite{lin2021arx:centerpose}  & 0.6649 & 0.6506 & 0.7780 & 0.7416 & 0.8361 & 0.8541 & 0.8428 & 0.6969 & 0.6631 & 0.7476 \\
None                       & 0.6810 & 0.6089 & 0.7734 & 0.7495 & 0.8491 & 0.8633 & 0.8488 & 0.7085 & 0.6566 & 0.7488 \\

\bottomrule
\end{tabular}
\end{table*}

\begin{figure}[t]
  \centering
  \vspace*{0.06in}
  \begin{tikzpicture}[inner sep = 0pt, outer sep = 0pt]
  
      \node[anchor=south west] (Book_GT) at (0in,0in)
      {\includegraphics[height=0.9in,clip=true,trim=0in 1.4in 0in 0.25in]{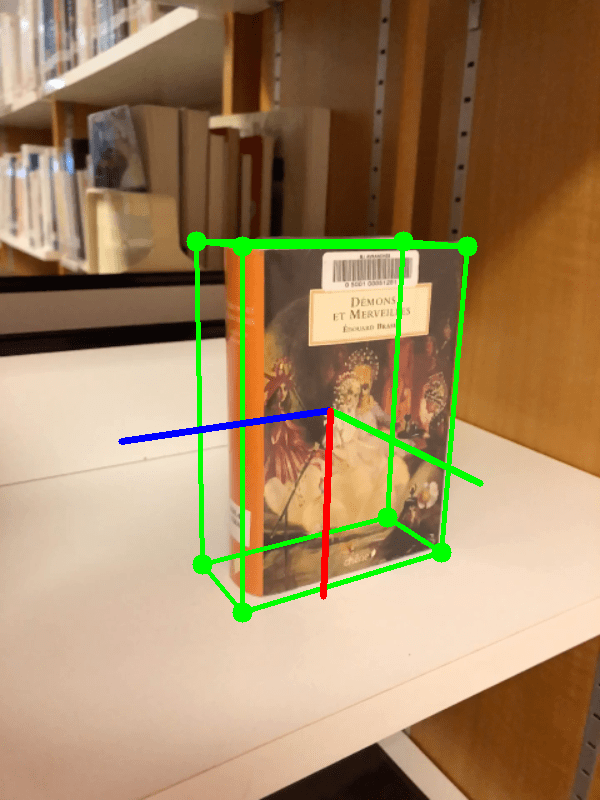}};
      \node[anchor=south,xshift=0pt,yshift=-10pt] at (Book_GT.south)
    {\small [3.33/1/2.43]};
      \node[anchor=south west,xshift=2pt] (Book_A) at (Book_GT.south east)
      {\includegraphics[height=0.9in,clip=true,trim=0in 1.4in 0in 0.25in]{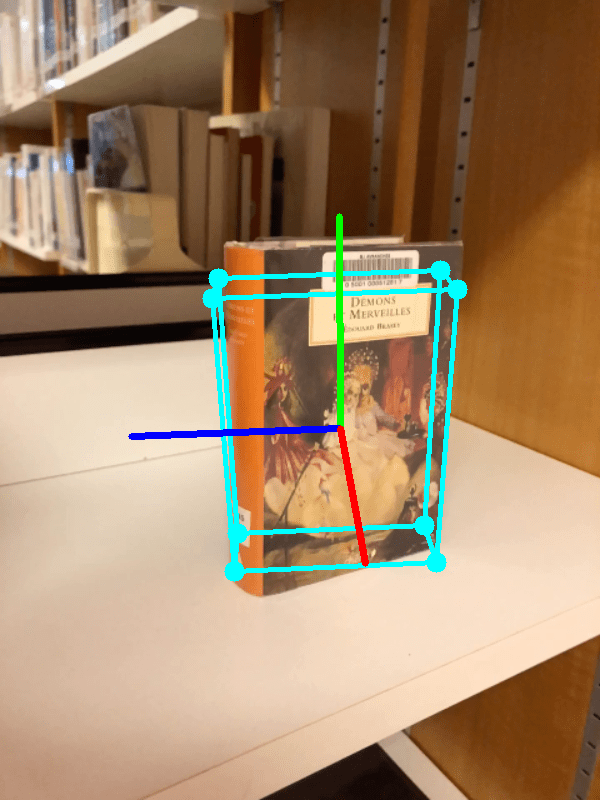}};
        \node[anchor=south,xshift=0pt,yshift=-10pt] at (Book_A.south)
        {\small [0.22/1/0.72]};
     \node[anchor=south west,xshift=2pt] (Book_B) at (Book_A.south east)
      {\includegraphics[height=0.9in,clip=true,trim=0in 1.4in 0in 0.25in]{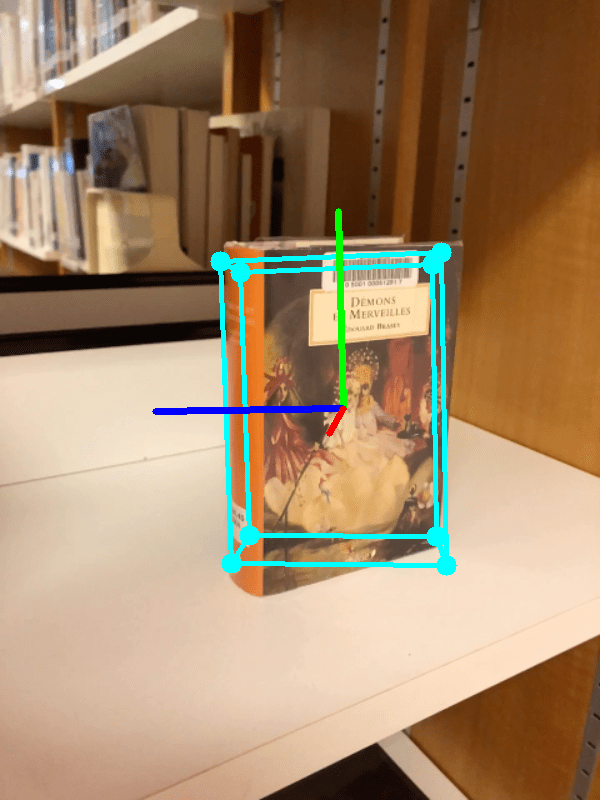}};
      \node[anchor=south,xshift=0pt,yshift=-10pt] at (Book_B.south)
        {\small [0.35/1/0.72]};
      \node[anchor=south west,xshift=2pt] (Book_C) at (Book_B.south east)
      {\includegraphics[height=0.9in,clip=true,trim=0in 1.4in 0in 0.25in]{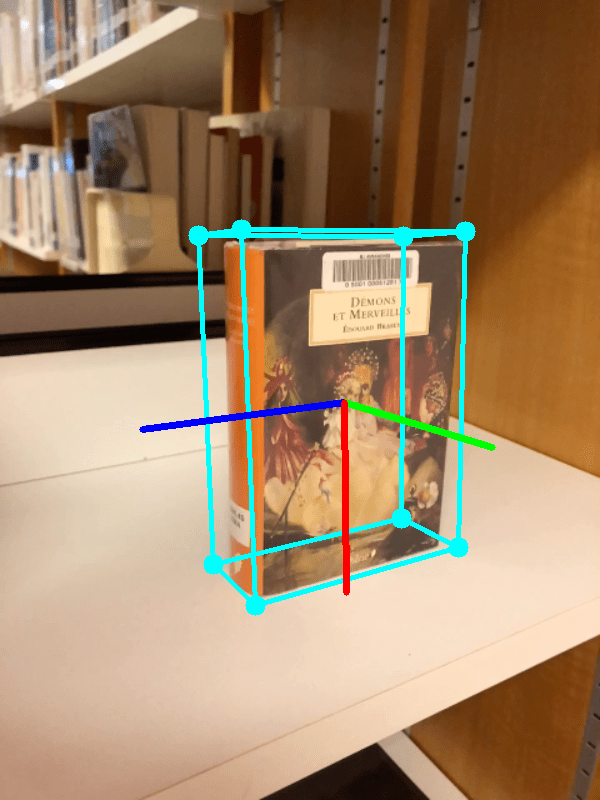}};
    \node[anchor=south,xshift=0pt,yshift=-10pt] at (Book_C.south)
        {\small [3.41/1/2.39]};
        
     \node[anchor=north west,yshift=-10pt] (Mug_GT) at (Book_GT.south west)
      {\includegraphics[height=0.9in,clip=true,trim=1.4in 3in 2.4in 3in]{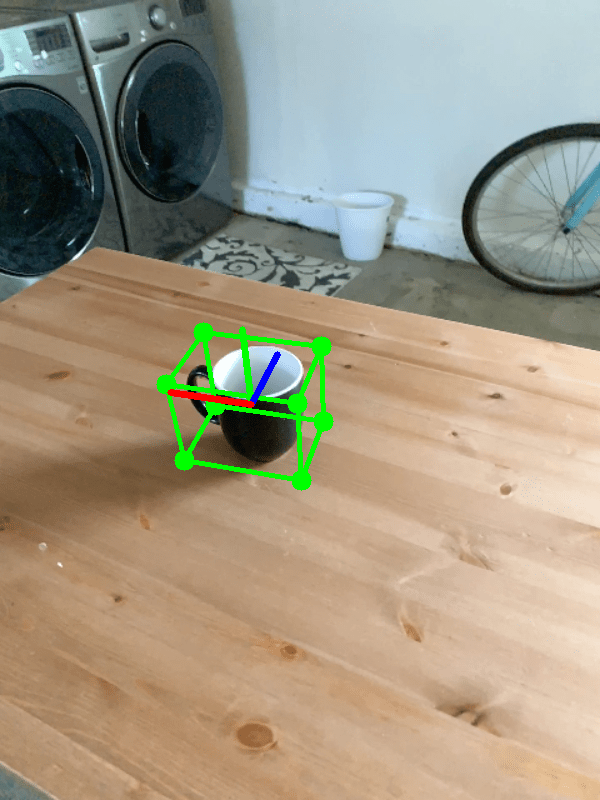}};
          \node[anchor=south,xshift=0pt,yshift=-10pt] at (Mug_GT.south)
        {\small [1.27/1/0.91]};
        
       \node[anchor=south west,xshift=2pt] (Mug_A) at (Mug_GT.south east)
      {\includegraphics[height=0.9in,clip=true,trim=1.4in 3in 2.4in 3in]{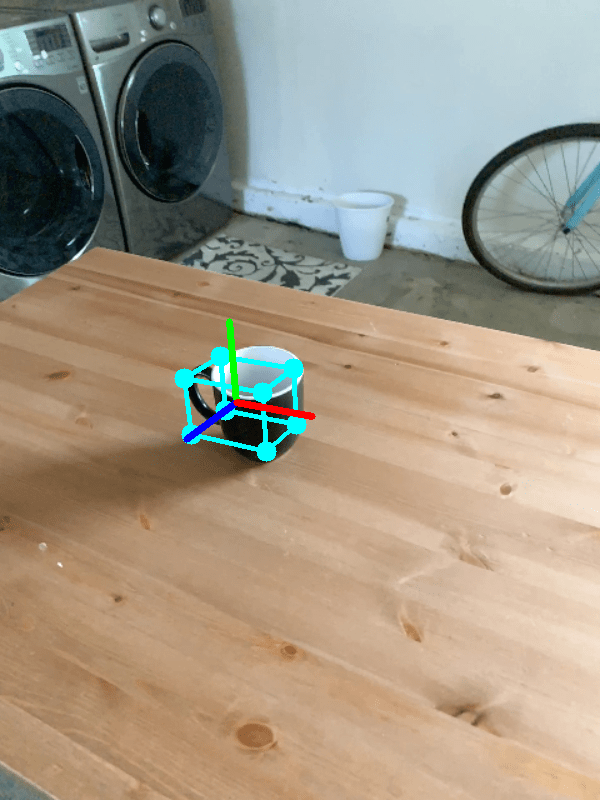}};
    \node[anchor=south,xshift=0pt,yshift=-10pt] at (Mug_A.south)
        {\small [1.40/1/1.00]};
        
     \node[anchor=south west,xshift=2pt] (Mug_B) at (Mug_A.south east)
      {\includegraphics[height=0.9in,clip=true,trim=1.4in 3in 2.4in 3in]{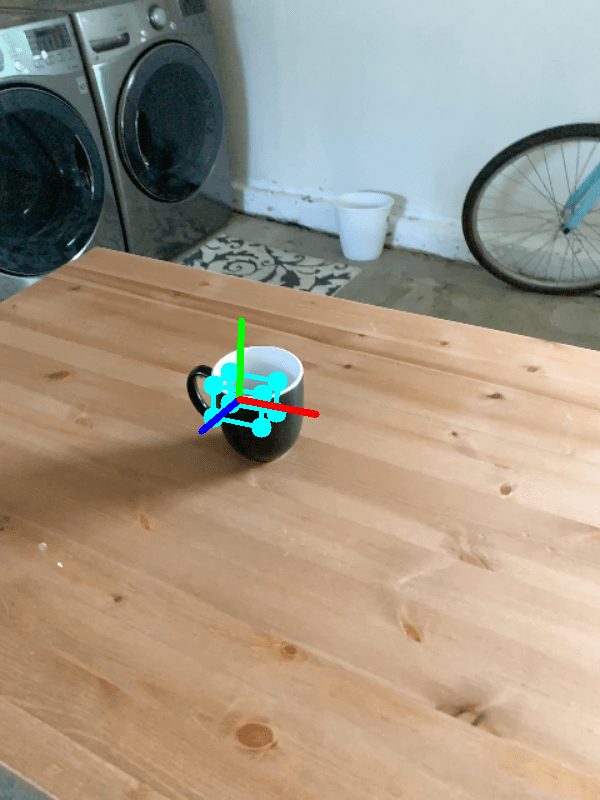}};
       \node[anchor=south,xshift=0pt,yshift=-10pt] at (Mug_B.south)
        {\small [1.5/1/1.11]};
      
      \node[anchor=south west,xshift=2pt] (Mug_C) at (Mug_B.south east)
      {\includegraphics[height=0.9in,clip=true,trim=1.4in 3in 2.4in 3in]{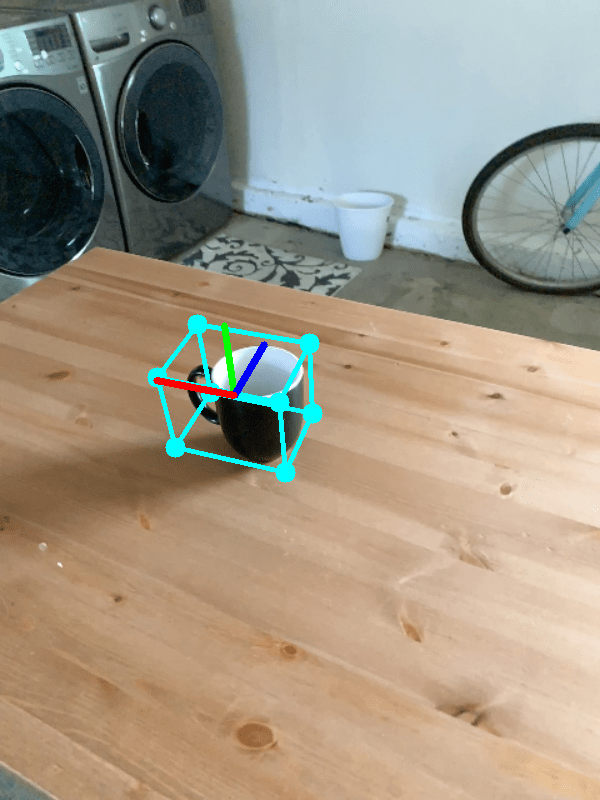}};
          \node[anchor=south,xshift=0pt,yshift=-10pt] at (Mug_C.south)
        {\small [1.30/1/0.95]};
    
    \node[anchor=south,xshift=0pt,yshift=-27pt] at (Mug_GT.south)
    {\small Ground truth};
    \node[anchor=south,xshift=0pt,yshift=-28.5pt] at (Mug_A.south)
    {\small CenterPose~\cite{lin2021arx:centerpose}};
   \node[anchor=south,xshift=0pt,yshift=-28.5pt] at (Mug_B.south)
    {\small w/ filtering};
\node[anchor=south,xshift=0pt,yshift=-27pt] at (Mug_C.south)
    {\small Ours};
  \end{tikzpicture}
  \caption{Qualitative comparison between CenterPose~\cite{lin2021arx:centerpose}, CenterPose w/ filtering, and our proposed method on book and cup images from Objectron~\cite{ahmadyan2021objectron}. The numbers show the relative dimensions of the 3D bounding box. In this case, simple filtering cannot recover from inaccurate predictions by the detector (Compare the second and third column).
  \label{fig:qualitative_comparison}} 
  \vspace*{-3 ex}
\end{figure}
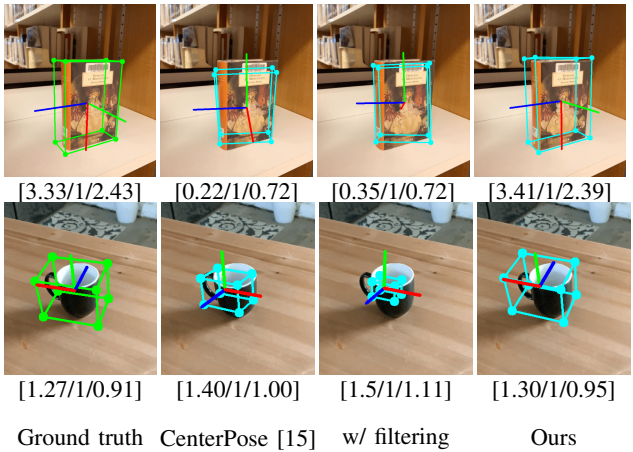

\subsection{Category-Level 6-DoF Pose and Size Tracking \label{exp:main}}

We compare the proposed method against 6 baselines including variants of our model: %
1)~{\it MobilePose}~\cite{hou2020mobilepose}: a single-stage light-weight detector with two heads regressing to the centroid location and the 3D bounding box keypoints.
2)~{\it Two-Stage}~\cite{ahmadyan2021objectron}: a two-stage detector for 2D object detection and 3D keypoint regression.
3)~{\it CenterPose}~\cite{lin2021arx:centerpose}: our prior single-stage detector.
4)~{\it CenterPose w/ filtering}~\cite{lin2021arx:centerpose}: our CenterPose detector extended with a simplified filtering process, where the uncertainty extracted from the heatmap is also used for the displacement output; and we directly average the relative dimension outputs across the frames in the fusion step.
5)~{\it Ours w/o filtering}: an ablation of our proposed method where the filtering process is removed.
6)~{\it Ours w/o heatmap}: an ablation of our proposed method where the prior heatmap input is empty.
7)~{\it Ours}: the complete version of our proposed method.
Similar to prior 6-DoF tracking works~\cite{wuthrich2013probabilistic, issac2016depth, wang20206}, in this experiment, we assume the ground truth pose and object size from the first frame are given to all the tracking methods.
We report the results in Table~\ref{table:main}, where the numbers for MobilePose~\cite{hou2020mobilepose} and Two-stage~\cite{ahmadyan2021objectron} are from~\cite{ahmadyan2021objectron}.

When compared with frame-by-frame detection methods~\cite{hou2020mobilepose,ahmadyan2021objectron,lin2021arx:centerpose}, our tracker achieves better results across all metrics. %
Our method improves by a large margin over the version without filtering, especially on the consistency metric, which suggests that the filtering process serves as a crucial component to explicitly alleviate the inter-frame jitter problem.
The version without heatmap inputs performs badly with occlusion (cup and shoe), which illustrates that our tracklet-conditioned mechanism provides a good prior region for the network to reason about, 
easing the difficulty when part of the object is not visible in the individual frame. 
 
\begin{figure}[t]
\centering
\begin{tabular}{rl|rl}
\includegraphics[trim=190 440 140 160, clip,width=0.20\columnwidth]{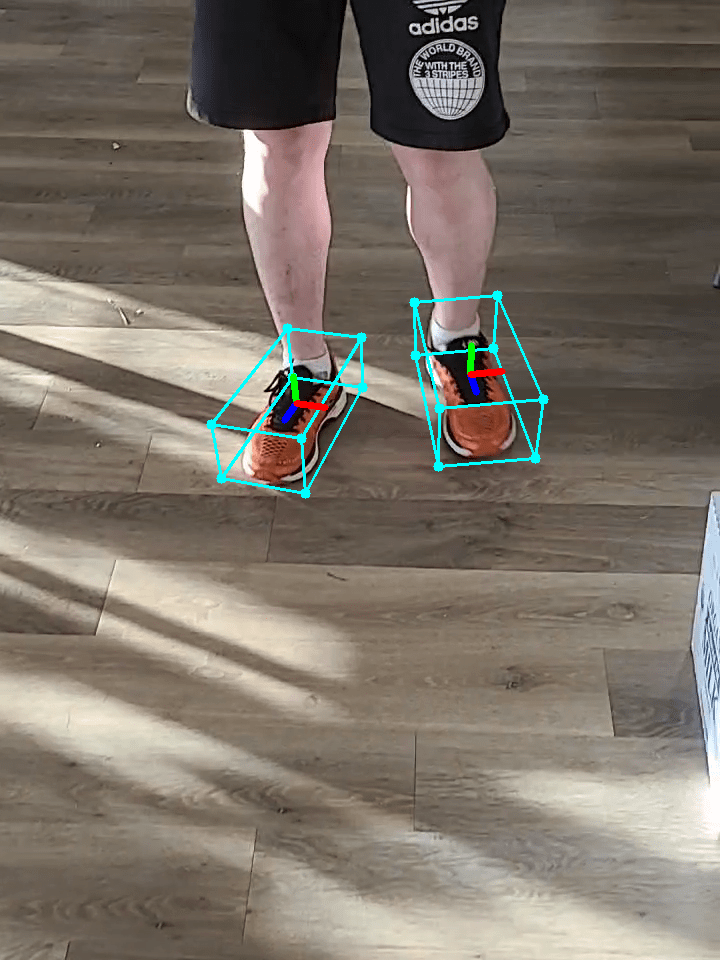} &
\includegraphics[trim=190 440 140 160, clip,width=0.20\columnwidth]{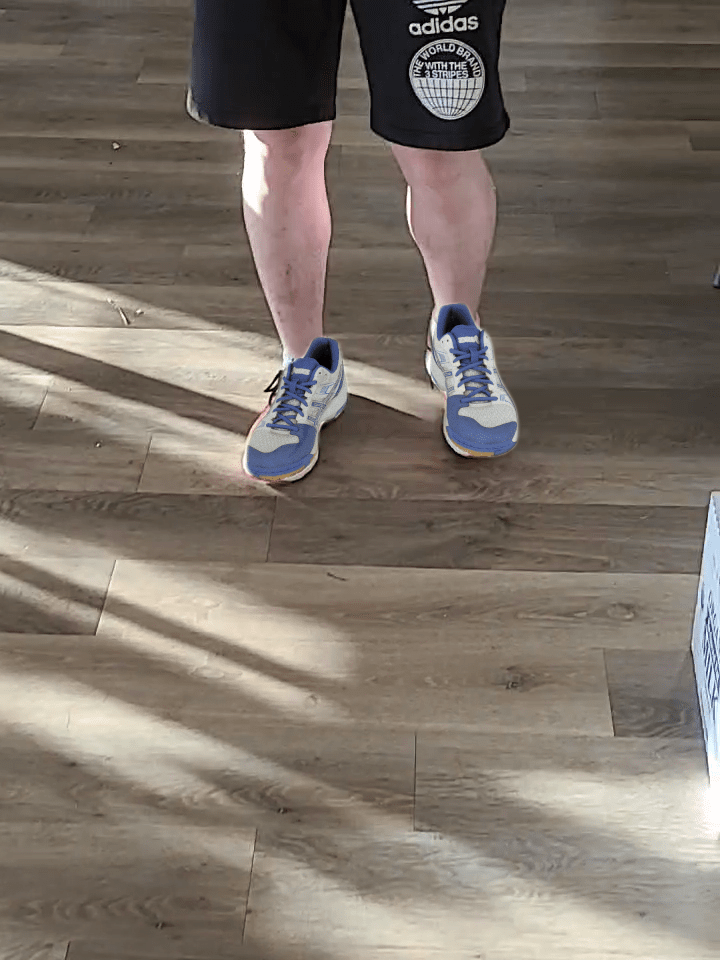} &
\includegraphics[trim=150 200 100 320, clip,width=0.20\columnwidth]{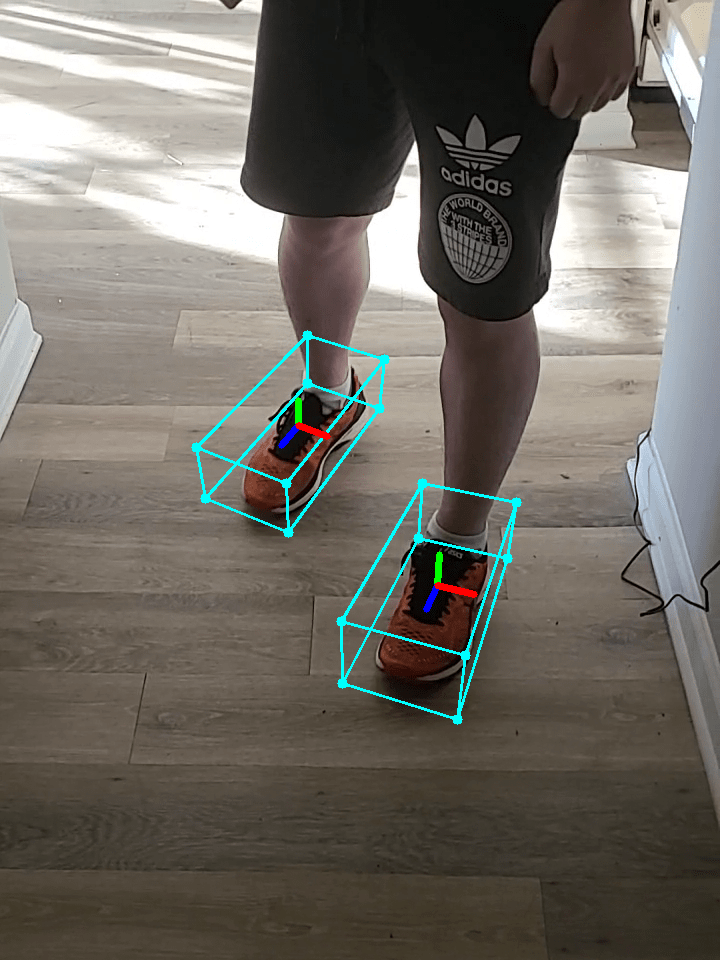} &
\includegraphics[trim=150 200 100 320, clip,width=0.20\columnwidth]{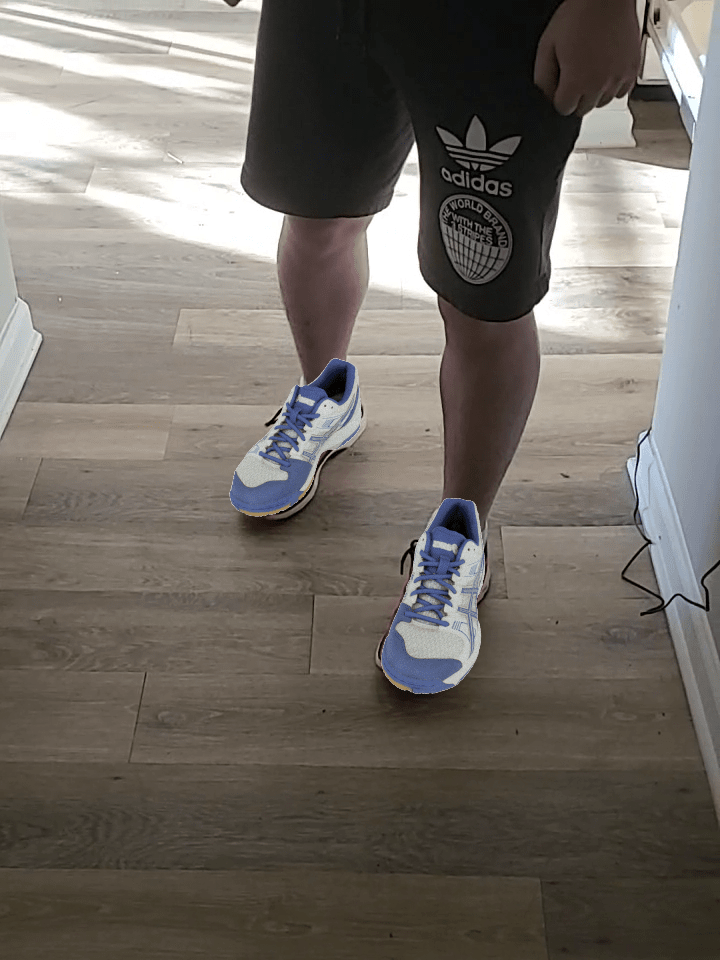} 
\end{tabular}
\caption{Frames from a real-world video sequence where CenterPoseTrack detected two shoes (left of each pair), 
and we overlay synthetic shoes (right of each pair) with matching poses. 
\label{fig:ar}} 
\vspace*{-3 ex}

\end{figure}

Figure~\ref{fig:qualitative_comparison} shows a qualitative comparison between CenterPose~\cite{lin2021arx:centerpose}, CenterPose w/ filtering, and our method.
For a real-world demonstration of the method, see the supplementary video which shows an augmented reality application in which poses and cuboid sizes are used to overlay a 3D model on a video, as previewed in Figure~\ref{fig:ar}.

\subsection{Pose Error under Different Track Initialization Methods}

Since our tracker is conditioned on the pose from the previous frame, it is important to test its robustness and stability with different initializations. 
We compare 4 initialization settings: 
1)~{\it Ground truth pose and size}: the same setting as used in Section~\ref{exp:main}.
2)~{\it Ground truth pose and size with noise}: following~\cite{weng2021captra}, we simulate Gaussian noise ($\sigma_{scale}$ = 20\%, $\sigma_{rot}$ = $5^{\circ}$, and $\sigma_{trans}$ = 3~cm) to mimic initialization error.
3)~{\it CenterPose}~\cite{lin2021arx:centerpose}: we use the prediction from our single-shot detector on the very first frame of each test video.
4)~{\it None}: since our proposed work is a joint detection and tracking approach, it can work without any external detector for initialization.

The results in Table~\ref{tab:init} highlight the robustness of our method to the initial pose, demonstrating that it can work independently of an external single-frame pose estimator.
Although, as expected, tracking performance decreases when ground truth is not available to initialize the first frame, 
accuracy of our method with no initialization (0.7488) is essentially the same as that obtained with CenterPose initialization (0.7476),
and is only slightly worse than that obtained when initialized with noisy ground truth (0.7734).
Moreover, it is important to note that even without initialization, CenterPoseTrack (0.7488) significantly outperforms the
frame-by-frame CenterPose detector (0.7218, see Table~\ref{table:main}).

\section{Conclusion}

We have proposed a joint detection  and  tracking method for category-level 6-DoF pose estimation of previously unseen object instances.
Our framework is designed to incorporate uncertainty estimation, both implicitly and explicitly, via a deep network and a filtering process. 
The proposed network works in a tracklet-conditioned manner by taking rendered heatmaps as input, and 
directly modeling prediction uncertainty in the outputs. %
To train the network, we also introduce a data generation process of simulating test-time errors.
Our proposed method achieves state-of-the-art performance on the Objectron benchmark.
Future work will aim to incorporate improved data encoding for the network and directly impose temporal consistency within a local region.

\bibliographystyle{IEEEtran}
\bibliography{main.bib}

\begin{thebibliography}{10}
\providecommand{\url}[1]{#1}
\csname url@samestyle\endcsname
\providecommand{\newblock}{\relax}
\providecommand{\bibinfo}[2]{#2}
\providecommand{\BIBentrySTDinterwordspacing}{\spaceskip=0pt\relax}
\providecommand{\BIBentryALTinterwordstretchfactor}{4}
\providecommand{\BIBentryALTinterwordspacing}{\spaceskip=\fontdimen2\font plus
\BIBentryALTinterwordstretchfactor\fontdimen3\font minus
  \fontdimen4\font\relax}
\providecommand{\BIBforeignlanguage}[2]{{%
\expandafter\ifx\csname l@#1\endcsname\relax
\typeout{** WARNING: IEEEtran.bst: No hyphenation pattern has been}%
\typeout{** loaded for the language `#1'. Using the pattern for}%
\typeout{** the default language instead.}%
\else
\language=\csname l@#1\endcsname
\fi
#2}}
\providecommand{\BIBdecl}{\relax}
\BIBdecl

\bibitem{pauwels2015simtrack}
K.~Pauwels and D.~Kragic, ``{SimTrack:} a simulation-based framework for
  scalable real-time object pose detection and tracking,'' in \emph{IEEE/RSJ
  International Conference on Intelligent Robots and Systems (IROS)}, 2015, pp.
  1300--1307.

\bibitem{tjaden2017real}
H.~Tjaden, U.~Schwanecke, and E.~Schomer, ``Real-time monocular pose estimation
  of {3D} objects using temporally consistent local color histograms,'' in
  \emph{Proceedings of the IEEE International Conference on Computer Vision
  (ICCV)}, 2017, pp. 124--132.

\bibitem{li2018deepim}
Y.~Li, G.~Wang, X.~Ji, Y.~Xiang, and D.~Fox, ``{DeepIM}: Deep iterative
  matching for {6D} pose estimation,'' in \emph{Proceedings of the European
  Conference on Computer Vision (ECCV)}, 2018, pp. 683--698.

\bibitem{wen2020se}
B.~Wen, C.~Mitash, B.~Ren, and K.~E. Bekris, ``{\textit{se}(3)-TrackNet}:
  Data-driven {6D} pose tracking by calibrating image residuals in synthetic
  domains,'' \emph{IEEE/RSJ International Conference on Intelligent Robots and
  Systems (IROS)}, 2020.

\bibitem{wang2019normalized}
H.~Wang, S.~Sridhar, J.~Huang, J.~Valentin, S.~Song, and L.~J. Guibas,
  ``Normalized object coordinate space for category-level {6D} object pose and
  size estimation,'' in \emph{Proceedings of the IEEE International Conference
  on Computer Vision (CVPR)}, 2019, pp. 2642--2651.

\bibitem{tian2020shape}
M.~Tian, M.~H. Ang, and G.~H. Lee, ``Shape prior deformation for categorical
  {6D} object pose and size estimation,'' in \emph{Proceedings of the European
  Conference on Computer Vision (ECCV)}, 2020, pp. 530--546.

\bibitem{chen2020learning}
D.~Chen, J.~Li, Z.~Wang, and K.~Xu, ``Learning canonical shape space for
  category-level {6D} object pose and size estimation,'' in \emph{Proceedings
  of the IEEE International Conference on Computer Vision (CVPR)}, 2020, pp.
  11\,973--11\,982.

\bibitem{chen2020category}
X.~Chen, Z.~Dong, J.~Song, A.~Geiger, and O.~Hilliges, ``Category level object
  pose estimation via neural analysis-by-synthesis,'' in \emph{Proceedings of
  the European Conference on Computer Vision (ECCV)}, 2020, pp. 139--156.

\bibitem{manhardt2020cps++}
F.~Manhardt, G.~Wang, B.~Busam, M.~Nickel, S.~Meier, L.~Minciullo, X.~Ji, and
  N.~Navab, ``{CPS++}: Improving class-level {6D} pose and shape estimation
  from monocular images with self-supervised learning,'' \emph{arXiv preprint
  arXiv:2003.05848}, 2020.

\bibitem{wang20206}
C.~Wang, R.~Mart{\'\i}n-Mart{\'\i}n, D.~Xu, J.~Lv, C.~Lu, L.~Fei-Fei,
  S.~Savarese, and Y.~Zhu, ``{6-PACK}: Category-level {6D} pose tracker with
  anchor-based keypoints,'' in \emph{IEEE International Conference on Robotics
  and Automation (ICRA)}, 2020, pp. 10\,059--10\,066.

\bibitem{weng2021captra}
Y.~Weng, H.~Wang, Q.~Zhou, Y.~Qin, Y.~Duan, Q.~Fan, B.~Chen, H.~Su, and L.~J.
  Guibas, ``{CAPTRA}: Category-level pose tracking for rigid and articulated
  objects from point clouds,'' \emph{arXiv preprint arXiv:2104.03437}, 2021.

\bibitem{feichtenhofer2017detect}
C.~Feichtenhofer, A.~Pinz, and A.~Zisserman, ``Detect to track and track to
  detect,'' in \emph{Proceedings of the IEEE International Conference on
  Computer Vision (ICCV)}, 2017, pp. 3038--3046.

\bibitem{zhang2018integrated}
Z.~Zhang, D.~Cheng, X.~Zhu, S.~Lin, and J.~Dai, ``Integrated object detection
  and tracking with tracklet-conditioned detection,'' \emph{arXiv preprint
  arXiv:1811.11167}, 2018.

\bibitem{bergmann2019tracking}
P.~Bergmann, T.~Meinhardt, and L.~Leal-Taixe, ``Tracking without bells and
  whistles,'' in \emph{Proceedings of the IEEE/CVF International Conference on
  Computer Vision (ICCV)}, 2019, pp. 941--951.

\bibitem{lin2021arx:centerpose}
Y.~Lin, J.~Tremblay, S.~Tyree, P.~A. Vela, and S.~Birchfield, ``Single-stage
  keypoint-based category-level object pose estimation from an {RGB} image,''
  in \emph{arXiv preprint arXiv:2109.06161}, 2021.

\bibitem{ahmadyan2021objectron}
A.~Ahmadyan, L.~Zhang, A.~Ablavatski, J.~Wei, and M.~Grundmann, ``Objectron: A
  large scale dataset of object-centric videos in the wild with pose
  annotations,'' in \emph{Proceedings of the IEEE International Conference on
  Computer Vision (CVPR)}, 2021, pp. 7822--7831.

\bibitem{zeng2017multi}
A.~Zeng, K.-T. Yu, S.~Song, D.~Suo, E.~Walker, A.~Rodriguez, and J.~Xiao,
  ``Multi-view self-supervised deep learning for {6D} pose estimation in the
  {A}mazon picking challenge,'' in \emph{IEEE International Conference on
  Robotics and Automation (ICRA)}, 2017, pp. 1386--1383.

\bibitem{sundermeyer2018implicit}
M.~Sundermeyer, Z.-C. Marton, M.~Durner, M.~Brucker, and R.~Triebel, ``Implicit
  {3D} orientation learning for {6D} object detection from {RGB} images,'' in
  \emph{Proceedings of the European Conference on Computer Vision (ECCV)},
  2018, pp. 699--715.

\bibitem{xiang2018posecnn}
Y.~Xiang, T.~Schmidt, V.~Narayanan, and D.~Fox, ``{PoseCNN}: A convolutional
  neural network for {6D} object pose estimation in cluttered scenes,'' in
  \emph{Robotics: Science and Systems (RSS)}, 2018.

\bibitem{rad2017bb8}
M.~Rad and V.~Lepetit, ``{BB8}: A scalable, accurate, robust to partial
  occlusion method for predicting the {3D} poses of challenging objects without
  using depth,'' in \emph{Proceedings of the IEEE International Conference on
  Computer Vision (ICCV)}, 2017, pp. 3828--3836.

\bibitem{tremblay2018deep}
J.~Tremblay, T.~To, B.~Sundaralingam, Y.~Xiang, D.~Fox, and S.~Birchfield,
  ``Deep object pose estimation for semantic robotic grasping of household
  objects,'' in \emph{Conference on Robot Learning (CoRL)}, 2018, pp. 306--316.

\bibitem{peng2019pvnet}
S.~Peng, Y.~Liu, Q.~Huang, X.~Zhou, and H.~Bao, ``{PVNet}: Pixel-wise voting
  network for 6{DoF} pose estimation,'' in \emph{Proceedings of the European
  Conference on Computer Vision (ECCV)}, 2019, pp. 4561--4570.

\bibitem{hou2020mobilepose}
T.~Hou, A.~Ahmadyan, L.~Zhang, J.~Wei, and M.~Grundmann, ``{MobilePose}:
  Real-time pose estimation for unseen objects with weak shape supervision,''
  \emph{arXiv preprint arXiv:2003.03522}, 2020.

\bibitem{choi2013rgb}
C.~Choi and H.~I. Christensen, ``{RGB-D} object tracking: A particle filter
  approach on {GPU},'' in \emph{IEEE/RSJ International Conference on
  Intelligent Robots and Systems (IROS)}, 2013, pp. 1084--1091.

\bibitem{wuthrich2013probabilistic}
M.~W{\"u}thrich, P.~Pastor, M.~Kalakrishnan, J.~Bohg, and S.~Schaal,
  ``Probabilistic object tracking using a range camera,'' in \emph{IEEE/RSJ
  International Conference on Intelligent Robots and Systems (IROS)}, 2013, pp.
  3195--3202.

\bibitem{deng2019pose}
X.~Deng, A.~Mousavian, Y.~Xiang, F.~Xia, T.~Bretl, and D.~Fox, ``{PoseRBPF}: A
  {Rao-Blackwellized} particle filter for {6D} object pose tracking,'' in
  \emph{Robotics: Science and Systems (RSS)}, 2019.

\bibitem{issac2016depth}
J.~Issac, M.~W{\"u}thrich, C.~G. Cifuentes, J.~Bohg, S.~Trimpe, and S.~Schaal,
  ``Depth-based object tracking using a robust gaussian filter,'' in \emph{IEEE
  International Conference on Robotics and Automation (ICRA)}, 2016, pp.
  608--615.

\bibitem{joseph2015versatile}
D.~Joseph~Tan, F.~Tombari, S.~Ilic, and N.~Navab, ``A versatile learning-based
  {3D} temporal tracker: Scalable, robust, online,'' in \emph{Proceedings of
  the IEEE International Conference on Computer Vision (ICCV)}, 2015, pp.
  693--701.

\bibitem{kendall2017uncertainties}
A.~Kendall and Y.~Gal, ``What uncertainties do we need in {B}ayesian deep
  learning for computer vision?'' in \emph{Conference on Neural Information
  Processing Systems (NIPS)}, 2017.

\bibitem{feng2020review}
D.~Feng, A.~Harakeh, S.~Waslander, and K.~Dietmayer, ``A review and comparative
  study on probabilistic object detection in autonomous driving,'' \emph{arXiv
  preprint arXiv:2011.10671}, 2020.

\bibitem{pan2020towards}
H.~Pan, Z.~Wang, W.~Zhan, and M.~Tomizuka, ``Towards better performance and
  more explainable uncertainty for {3D} object detection of autonomous
  vehicles,'' in \emph{IEEE International Conference on Intelligent
  Transportation Systems (ITSC)}, 2020.

\bibitem{he2020deep}
Y.~He and J.~Wang, ``Deep mixture density network for probabilistic object
  detection,'' in \emph{IEEE/RSJ International Conference on Intelligent Robots
  and Systems (IROS)}, 2020.

\bibitem{harakeh2020bayesod}
A.~Harakeh, M.~Smart, and S.~L. Waslander, ``{BayesOD}: A {B}ayesian approach
  for uncertainty estimation in deep object detectors,'' in \emph{IEEE
  International Conference on Robotics and Automation (ICRA)}, 2020, pp.
  87--93.

\bibitem{dong2020probabilistic}
X.~Dong, P.~Wang, P.~Zhang, and L.~Liu, ``Probabilistic oriented object
  detection in automotive radar,'' in \emph{IEEE/CVF Conference on Computer
  Vision and Pattern Recognition Workshops (CVPRW)}, 2020, pp. 102--103.

\bibitem{meyer2020learning}
G.~P. Meyer and N.~Thakurdesai, ``Learning an uncertainty-aware object detector
  for autonomous driving,'' in \emph{IEEE/RSJ International Conference on
  Intelligent Robots and Systems (IROS)}, 2020, pp. 10\,521--10\,527.

\bibitem{zhou2020tracking}
X.~Zhou, V.~Koltun, and P.~Kr{\"a}henb{\"u}hl, ``Tracking objects as points,''
  in \emph{Proceedings of the European Conference on Computer Vision (ECCV)},
  2020, pp. 474--490.

\bibitem{tremblay2018synthetically}
J.~Tremblay, T.~To, A.~Molchanov, S.~Tyree, J.~Kautz, and S.~Birchfield,
  ``Synthetically trained neural networks for learning human-readable plans
  from real-world demonstrations,'' in \emph{IEEE International Conference on
  Robotics and Automation (ICRA)}, 2018, pp. 5659--5666.

\bibitem{kraus2019uncertainty}
F.~Kraus and K.~Dietmayer, ``Uncertainty estimation in one-stage object
  detection,'' in \emph{IEEE Intelligent Transportation Systems Conference
  (ITSC)}, 2019, pp. 53--60.

\bibitem{kumar2006generalized}
M.~Kumar, D.~P. Garg, and R.~A. Zachery, ``A generalized approach for
  inconsistency detection in data fusion from multiple sensors,'' in
  \emph{American Control Conference (ACC)}, 2006, pp. 2078--2083.

\bibitem{abdel2015direct}
Y.~I. Abdel-Aziz, H.~M. Karara, and M.~Hauck, ``Direct linear transformation
  from comparator coordinates into object space coordinates in close-range
  photogrammetry,'' \emph{Photogrammetric Engineering \& Remote Sensing},
  vol.~81, no.~2, pp. 103--107, 2015.

\end{thebibliography}

\end{document}